\documentclass[runningheads]{llncs}
\usepackage{graphicx}
\usepackage{subcaption}
\usepackage{hyperref}
\begin{document}
\title{Traffic sign detection and recognition using event camera image reconstruction}

\titlerunning{Traffic sign detection and recognition using event cameras}
%
\author{Kamil Jeziorek \href{https://orcid.org/0000-0001-5446-3682}{\includegraphics[width=16pt]{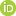}} \and
Tomasz Kryjak \href{https://orcid.org/0000-0001-6798-4444}{\includegraphics[width=16pt]{orcid.png}} }
\authorrunning{K. Jeziorek, T. Kryjak}
%
\institute{AGH University of Science and Technology, \\ Al. Mickiewicza 30, 30-059 Krakow, Poland\\
\email{kjeziorek@student.agh.edu.pl, tomasz.kryjak@agh.edu.pl} \\
}

\maketitle              
\begin{abstract}
This paper presents a~method for detection and recognition of traffic signs based on information extracted from an event camera. 
The solution used a~FireNet deep convolutional neural network to reconstruct events into greyscale frames. 
Two YOLOv4 network models were trained, one based on greyscale images and the other on colour images. 
The best result was achieved for the model trained on the basis of greyscale images, achieving an efficiency of 87.03\%.

\end{abstract}


\section{Introduction}

In modern cars, safety and driving comfort are key aspects. Therefore, manufacturers are equipping them with so-called Advanced Driver Assistance Systems (ADAS). These systems use a~range of sensors, including video and thermal cameras, radars, ultrasonics, and sometimes LiDARs (Light Detection and Ranging). Their task is to recognise traffic signs, stay in the lane, and detect other vehicles, pedestrians, and cyclists. The standard frame cameras currently in use have significant drawbacks, such as low dynamic range, which makes it difficult to work properly in difficult lighting conditions; low temporal resolution, which prevents the acquisition of relevant information between frames; and the blurring effect that can occur when the environment changes dynamically or the camera moves (especially in low-light situations). One possible solution to the mentioned disadvantages is a~modern event camera (also called a dynamic vision sensor (DVS) or a neuromorphic camera), based on the principle of the human eye.

In this work, a~computer vision system was designed to recognise traffic signs, based on information from an event camera. Due to its novel nature, there is still no sufficiently developed algorithm base for object detection and classification. For this reason, a~deep neural network was used to reconstruct the information from the event camera into greyscale frames. This allowed us to use an algorithm designed to process images acquired from a~traditional video camera.

The remainder of this article is organised as follows. Section \ref{sec:event} describes the principle of the event camera and its advantages over the traditional frame camera. Section \ref{sec:event_reconstruction} lists the possible representations of information extracted from the event camera along with the neural networks responsible for their reconstruction. Section \ref{sec:proposed_system} presents the realised computer vision system and the temporal and qualitative results obtained from the algorithms used. The article concludes with a~summary and an indication of the potential application of the solution.

\section{Event camera}
\label{sec:event}

\begin{figure}[!t]
\centering
\includegraphics[width=\textwidth]{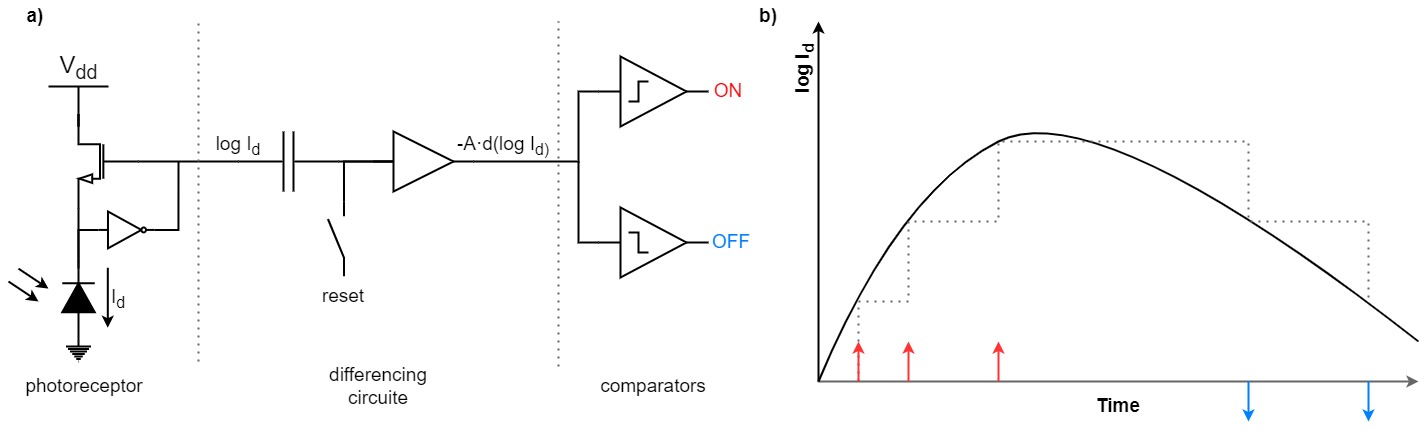}
\caption{The principle of the pixel in an event camera: a) Simplified diagram of the pixel. The figure shows 3 main components: A photosensitive element, a~module for determining the difference in voltage from the initial value with a~reset input, and a~module composed of two comparators that determine the type of change in light intensity, b) A graph showing the generation of an event. Based on: \cite{2}.} 
\label{fig:1}
\end{figure}

Event cameras are modern sensors inspired by the principle of the human eye. Each pixel in the sensor operates independently of the rest and continuously responds to changes in ambient light intensity. A~simplified diagram of a~pixel is shown in Figure \ref{fig:1}a. It stores the logarithm of light intensity, and when the threshold value is exceeded, an event is generated, and the stored intensity value is reset (Figure \ref{fig:1}b.). Each event consists of four pieces of information: the time at which the information was generated (timestamp), the position of the pixel on the matrix (x and y, respectively), and the polarity, which determines whether the change in illumination is positive or negative from the reference value.

Since each pixel operates independently of each other and responds to changes in light intensity, information is sent through the camera only at times when the scene changes accordingly. This results in lower power consumption compared to a~standard camera (in a~typical situation). Also, the high temporal resolution of the event sensors (1 million events per second) allows recording very fast movement. In contrast, a traditional video camera takes information from each pixel simultaneously at a~specific frequency. In order to obtain light intensity information, the pixels need an appropriate exposure time, which, when recording a~dynamically changing environment, can result in blurred images.
Another important difference between the two cameras is the dynamic (tonal) range. This parameter determines the ability to record extremely differently exposed points in the scene. The larger the tonal range, the greater the difference between two points can be. Standard cameras reach values of about 50-60 dB, when event cameras can reach up to 120 dB. This allows to achieve much better results in situations where scenes are poorly illuminated, or the difference between the brightest and darkest values is very large. Examples of such scenarios include recordings during the night or when the sun is in the camera's field of view, or entering/exiting a~tunnel. 

\section{Event reconstruction}
\label{sec:event_reconstruction}

Event cameras have a~number of advantages, which are described in detail in Section \ref{sec:event}. However, acting on an asynchronous sequence of events is not a~simple task. To be able to use the information obtained from an event camera, various forms of event representation are used. Among them, the event frame and the 3D point cloud can be distinguished. The first form projects the obtained events onto a~2D plane. This is a~simple operation, but the disadvantage is the loss of information about the time of occurrence of individual events. The second representation uses the time of occurrence of an event and appropriately represents it in 3D space. This is a~more complicated representation, but the advantage is the preservation of timing information.
The solutions discussed have made it easier to work on the events themselves, but they are still not compatible with the image processing algorithms developed so far. Another possibility is to reconstruct standard images based on a~sequence of events while retaining the advantages of an event camera. The first attempts used an Extended Kalman Filter or a~complementary filter for light intensity estimation. However, more recent approaches have focused on using fully convolutional neural networks, which allowed to obtain better results.

\begin{figure}[!t]
\includegraphics[width=\textwidth]{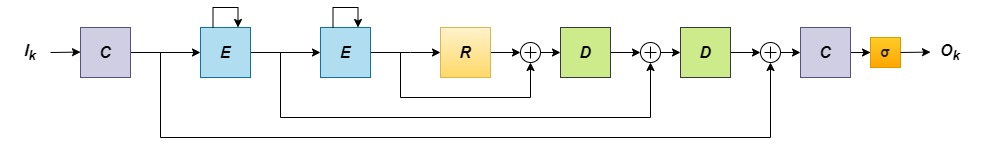}
\caption{E2VID network architecture. The input and output convolution layers are shown in purple, the encoders are shown in blue, the residual block in yellow, and the decoders in green. Based on: \cite{6}} 
\label{fig:2}
\end{figure}

The first example of an algorithm using deep learning for reconstruction is the E2VID model \cite{6}. It is a~network inspired by a~U-NET type architecture. It is characterised by a~symmetric structure, which is shown in \ref{fig:2}. It consists of 7~layers: one convolutional input layer, then two encoders, composed of a~convolutional layer and a~recursive connection. The output of the encoders produces reduced images, due to the result of the convolution operation. The middle part consists of a~residual block, which has additional connections between layers. The next two modules are decoders, which are designed to reconstruct the initial resolution of the image. At the output of the last layer, the prediction is performed. Each symmetric block has additional connections that skip the internal blocks.
The second example is FireNet \cite{7}, which is a~streamlined version of the E2VID model. It consists of 6 blocks, and the layers that modify the image size have been dropped. Also a~faster and simpler encoder design has been used. As a~result, there are as many as 280 times fewer parameters to learn and several times shorter computation execution times compared to the E2VID network. Due to this fact, we used FireNet in our experiments.

\section{Implementation of the computer vision system}
\label{sec:proposed_system}

The algorithm responsible for the detection and recognition of traffic signs consisted of three main components:

\begin{enumerate}
    \item an event camera data source, stored in \textit{txt} format,
    \item the network responsible for reconstructing the events into greyscale frames,
    \item the algorithm responsible for detection and recognition of traffic signs based on the reconstructed images.
\end{enumerate}

In these experiments, the resources of the NVIDIA GeForce RTX 2060 graphics card were utilised.
In this work, the DSEC database \cite{3} recordings and Driving Event Camera Datasets \cite{5} were used. 
Both collections were made with an event camera with a~resolution of 640x480 pixels and reference recordings were made with standard RGB cameras. 
Using the recordings from conventional cameras, interesting moments were selected -- when a~car passed traffic signs. Knowing the time intervals of the sign occur, an appropriate number of events were cut from the file. The data was then converted to \textit{txt} text format, which is required for reconstruction.

\begin{table}[!t]
\caption{Image reconstruction time depending on the event retrieval mode and the number of events.}
\centering
\resizebox{\textwidth}{!}{%
\begin{tabular}{|c|c|c|c|c|}
\cline{1-2} \cline{4-5}
\textbf{Event retrieval time [ms]} & \textbf{Reconstruction time [ms]} & \phantom{.....} & \textbf{Number of events} & \textbf{Reconstruction time [ms]} \\ \cline{1-2} \cline{4-5} 
1                         & 19.15                    &                          & 10000            & 22.22                    \\ \cline{1-2} \cline{4-5} 
2                         & 25.09                    &                          & 20000            & 30.11                    \\ \cline{1-2} \cline{4-5} 
3                         & 28.18                    &                          & 30000            & 41.48                    \\ \cline{1-2} \cline{4-5} 
4                         & 37.50                    &                          & 40000            & 43.55                    \\ \cline{1-2} \cline{4-5} 
5                         & 40.77                    &                          & 50000            & 45.32                    \\ \cline{1-2} \cline{4-5} 
6                         & 42.95                    &                          & 60000            & 50.86                    \\ \cline{1-2} \cline{4-5} 
7                         & 45.00                    &                          & 70000            & 53.49                    \\ \cline{1-2} \cline{4-5} 
8                         & 47.74                    &                          & 80000            & 56.43                    \\ \cline{1-2} \cline{4-5} 
9                         & 51.12                    &                          & 90000            & 59.85                    \\ \cline{1-2} \cline{4-5} 
10                        & 52.25                    &                          & 100000           & 64.06                    \\ \cline{1-2} \cline{4-5} 
\end{tabular}%
}
\label{table:1}
\end{table}

FireNet was used for image reconstruction. To perform the procedure, it is enough to pass the created text file, containing information about the events, to the input of the network. There are several options that can be set during reconstruction. The most important of these is the selection of the event retrieval mode. One can set whether the number of events to generate one frame will be fixed or the time at which events are collected will be fixed. Both forms have their advantages and disadvantages. When a~fixed number of events is selected, each frame will be calculated with the same time and accuracy. However, in this case, the generated frames have no reference to the time when the events occurred, and if the number of generated events is very low then collecting a~sufficient number for reconstruction requires a~long waiting time. In contrast, collecting events at a~fixed time guarantees that each frame will be created at the same time, which makes it possible to predict how many frames will be generated in one second. However, it is not known how many events the camera has generated in a~given period of time, so the generation time can vary. The time required to reconstruct one frame depending on the number of events downloaded and the mode is shown in Table \ref{table:1}. 

The YOLOv4 network \cite{1} was used for the detection and recognition of traffic signs due to its high accuracy and rather low computational demand. To train the network to recognise traffic signs, a~training set consisting of images and annotations specifying the location and class of signs had been prepared. The GTSDB \cite{4} (German Traffic Sign Detection Benchmark) database was chosen. It consists of 900 photos with a~size of 1360x800 and a~text file with specific locations of bounding boxes and sign classes. Each marked sign is assigned to one of the 43 classes. However, it was decided to reduce them to 4 general groups: prohibition, mandatory, warning, and other. The goal was to improve the distribution of occurrences of each sign per class. The learning was done using the transfer learning method. This allows the use of previously trained network weights, for the new task of traffic sign class detection. 

\begin{table}[!t]
\caption{Quality indicators of both trained models for each class separately. The AP index values obtained for the metrics mAP@0.5IoU and mAP@0.75IoU are shown.}
\resizebox{\textwidth}{!}{%
\begin{tabular}{|c|ll|ll|}
\hline
\multicolumn{1}{|l|}{} & \multicolumn{2}{c|}{Greyscale} & \multicolumn{2}{c|}{Color} \\ \hline
Sign class & \multicolumn{1}{l|}{mAP@0.5IoU {[}\%{]}} & mAP@0.75IoU {[}\%{]} & \multicolumn{1}{l|}{mAP@0.5IoU {[}\%{]}} & mAP@0.75IoU {[}\%{]} \\ \hline
Prohibition & \multicolumn{1}{l|}{77.78} & 50 & \multicolumn{1}{l|}{60} & 30 \\ \hline
Warning & \multicolumn{1}{l|}{100} & 50 & \multicolumn{1}{l|}{100} & 0 \\ \hline
Mandatory & \multicolumn{1}{l|}{76} & 60 & \multicolumn{1}{l|}{20} & 20 \\ \hline
Other & \multicolumn{1}{l|}{94.36} & 38.89 & \multicolumn{1}{l|}{76.67} & 35.56 \\ \hline
\end{tabular}%
}
\label{table:2}
\end{table}

\begin{figure}[!t]
\begin{subfigure}{0.32\textwidth}
\includegraphics[width=\textwidth]{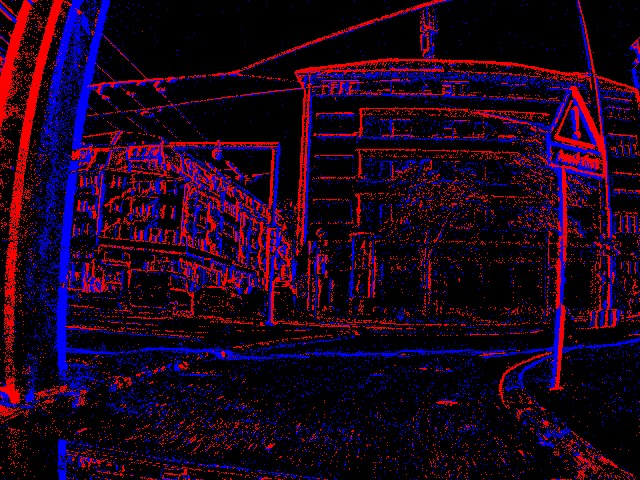}
\caption{Used events.} 
\label{fig:3a}
\end{subfigure}
\begin{subfigure}{0.32\textwidth}
\includegraphics[width=\textwidth]{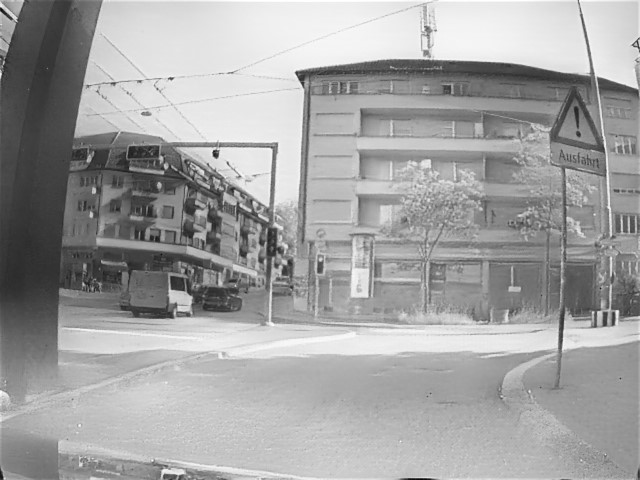}
\caption{Reconstruction.} 
\label{fig:3b}
\end{subfigure}
\begin{subfigure}{0.32\textwidth}
\includegraphics[width=\textwidth]{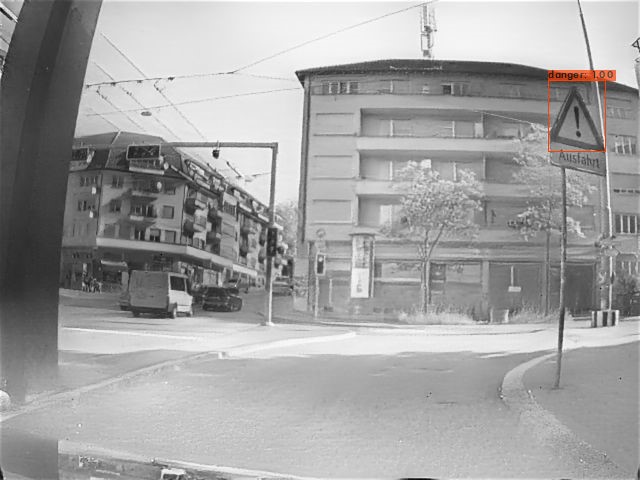}
\caption{Detection.} 
\label{fig:3c}
\end{subfigure}
\caption{An example of the result of each stage of the proposed computer vision system.} 
\label{fig:3}
\end{figure}

Two separate models were trained: based on original colour images and images converted to greyscale. Qualitative tests were carried out on a~test base, prepared on the basis of the images obtained during reconstruction. The results are shown in Table \ref{table:2}, while the effects of reconstruction and detection are shown in Figure \ref{fig:3}.

For detection, using a~threshold value of 0.5, the algorithm performs best in detecting warning signs, followed by a~class of other signs. Both classes have distinctive shapes. In contrast, the model learnt from colour images performs less well at detecting prohibition and mandatory signs, which share the same shape and are distinguished by the colour used. Increasing the threshold to 0.75 significantly affects the detection of all signs, especially warning signs. As a~result, the model learnt from greyscale images proved to be a~better detector, achieving a~high overall mAP@0.5IoU rate of 87.03\%.

FireNet proved to be computationally complex. The time required to perform reconstruction based on the events collected at 1 ms and 10 ms was 19.15 ms and 52.25 ms, respectively. The values obtained are too high to implement the algorithm in real time.

\section{Summary}

The result of the work described here is a~computer vision system that enables traffic sign detection based on event camera data. The tests carried out show that recognition of traffic signs is possible. The problem turned out to be the performance of the reconstruction operation.

The solution to the problem may be to use only part of the information, at the expense of reconstruction quality. Another way to improve the speed of operations may be to use a~different computing platform that allows faster and parallel calculations. However, the current version of the algorithm can be used to generate a new training set based on event information, which can be then used to train algorithms that are time-optimised and adapted to the event format.

%
%
%
%

\end{document}